\pdfoutput=1

\documentclass[11pt]{article}

\usepackage[preprint]{acl}

\usepackage{times}
\usepackage{latexsym}
\usepackage{amsfonts}
\usepackage{float}

\usepackage[T1]{fontenc}

\usepackage[utf8]{inputenc}

\usepackage{microtype}

\usepackage{inconsolata}

\usepackage{graphicx}
\usepackage{amsmath}
\usepackage{booktabs}
\usepackage{multirow}
\usepackage{xurl}

\title{GLiDRE: Generalist Lightweight model for Document-level Relation Extraction}

\author{
  Robin Armingaud, Romaric Besançon \\
  Université Paris-Saclay, CEA, List, F-91120, Palaiseau, France \\
  \texttt{\{robin.armingaud, romaric.besancon\}@cea.fr}
}

\begin{document}
\maketitle
\begin{abstract}

Relation Extraction (RE) is a fundamental task in Natural Language Processing, and its document-level variant poses significant challenges, due to complex interactions between entities across sentences. While supervised models have achieved strong results in fully resourced settings, their behavior with limited training data remains insufficiently studied. We introduce GLiDRE, a new compact model for document-level relation extraction, designed to work efficiently in both supervised and few-shot settings. Experiments in both low-resource supervised training and few-shot meta-learning benchmarks show that our approach outperforms existing methods in data-constrained scenarios, establishing a new state-of-the-art in few-shot document-level relation extraction. Our code will be publicly available\footnote{\url{https://github.com/robinarmingaud/glidre}}.
\end{abstract}

\section{Introduction}

Document-level Relation Extraction (DocRE) is a challenging task in natural language processing that involves identifying relationships between entities across multiple sentences. In contrast, traditional sentence-level relation extraction, evaluated on supervised datasets like TACRED \citep{zhang-etal-2017-position} and SemEval-2010 Task 8 \citep{hendrickx-etal-2010-semeval} or few-shot benchmarks such as FewREL \citep{han-etal-2018-fewrel} and Wiki-ZSL \citep{chen-li-2021-zs}, typically considers a single entity pair per instance, often without negative examples (i.e. sentences where no relation is present). These settings closely resemble relation classification rather than relation extraction. While recent works improve existing benchmarks, like FewRel 2.0 \citep{gao-etal-2019-fewrel} which addresses the negative examples issue, they still operate under the constraint of classifying a fixed pair of entities.

DocRE presents a more realistic and complex scenario, especially relevant for real-world applications such as biomedical or industrial Information Extraction, where relationships often span sentence boundaries and the entity pairs of interest are not pre-identified. The complexity of DocRE is due to the quadratic growth of negative pairs with the number of entities in a document. Evaluation is commonly conducted on datasets like DocRED or Re-DocRED \citep{yao-etal-2019-docred, tan-etal-2022-revisiting}. 

\begin{figure*}[t]
  \includegraphics[width=1\linewidth]{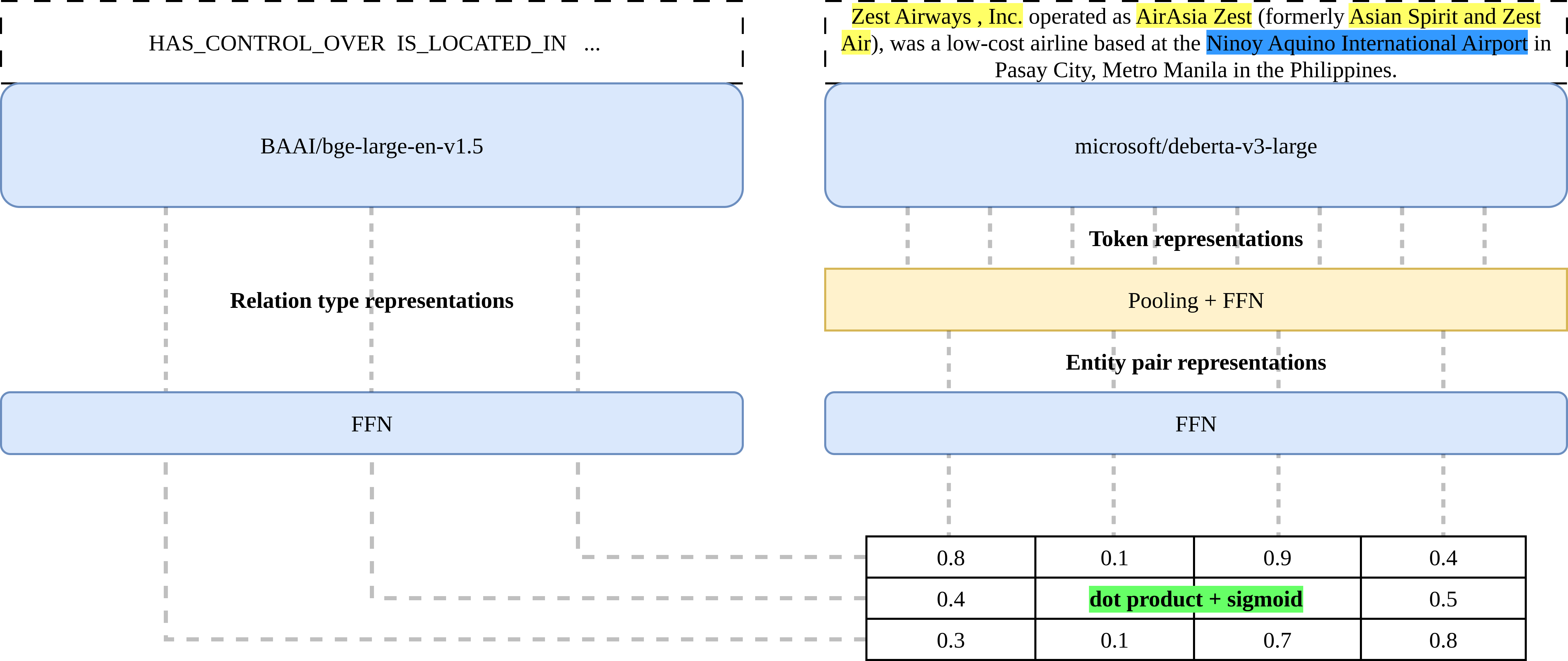}
  \caption {Model architecture. GLiDRE uses a dual-encoder structure: a text encoder generates token embeddings, while a label encoder computes embeddings for relation types. An entity representation is formed by pooling the embeddings of its constituent tokens and mentions. For each entity pair, their representations are concatenated and processed by a feed-forward network before final scoring. The model can optionally incorporate a localized context pooling module~\citep{zhou2021atlop} to enrich the relation representation.}
  \label{fig:glidre_model}
\end{figure*}

However, if recent works tackle sentence-level relation extraction in low-resource settings \citep{boylan-etal-2025-glirel,ijcai2024p702,MC-BERT-Lan-et-al}, there has been little research on DocRE in similar settings. In zero-shot settings, LLMs show strong capabilities for tasks such as NER and RE \citep{sainz2024gollie,zhou2023universalner,wang2023instructuie,wei2023zero}, but their performance on zero-shot DocRE is still limited
(\citealp{li-etal-2023-semi}, \citealp{xue-etal-2024-autore}). 
On the other hand, lightweight zero-shot models have also achieved great performances for NER, such as GLiNER \citep{zaratiana-etal-2024-gliner}, which exploits similarities between
entity type representations and textual span representations in the latent space. This approach allows better representations than LLMs due to its use of bidirectional encoders and it solves the scalability issues of autoregressive models, surpassing much larger LLMs in zero-shot settings. Moreover, as a zero-shot model, GLiNER is not restricted to a fixed set of relation labels and can generalize to arbitrary relation types at inference time. With the same inspiration, we offer the following contributions:
\begin{itemize}
    \item We propose GLiDRE, a new and efficient model for document-level relation extraction with zero or few annotated data, that leverages embedding similarities between relation types and entity pair representations.
    \item We conduct extensive experiments on the Re-DocRED, FREDo and Re-FREDo benchmarks, demonstrating that GLiDRE achieves state-of-the-art results in challenging few-shot scenarios.
    \item We demonstrate that smaller, specialized encoder models can offer an efficient alternative to larger LLMs for complex, document-level relation extraction and we present an up-to-date evaluation of these models on the Re-DocRED dataset in a zero-shot setting.
\end{itemize}

\section{Related Work}

\paragraph{Document-level RE methods}
State-of-the-art supervised approaches to DocRE generally extend the model ATLOP \citep{zhou2021atlop}, which introduces Localized Context Pooling, using attention mechanisms to enhance relation representations, and an adaptive-thresholding loss that adjusts decision thresholds dynamically based on entity pairs. Subsequent work has extended ATLOP in various directions. For example, DREEAM \citep{ma-etal-2023-dreeam} uses evidences to enhance performance, while KD-DocRE \citep{tan-etal-2022-document} uses axial attention to model interdependencies among entity pairs, adopts an adaptive focal loss to address class imbalance, and integrates knowledge distillation.
Other works enhance DocRE models with rule constraints, adding logical constraints to improve consistency, with techniques such as bidirectional rule mining \cite{liu-etal-2023-document-level} or differentiable document-specific rule learning \cite{zhang-etal-2025-cadrl}. Another trend leverages LLMs: for instance, LMRC \cite{li2024llm} combines a dedicated classifier to suggest candidate relations with the generative ability of a LLM for final decisions, significantly improving upon purely generative approaches.

\paragraph{Benchmarks for DocRE}
Document-level Relation Extraction models are commonly evaluated on the benchmark DocRED \cite{zhou2021atlop}, and its refined version Re-DocRED \citep{tan-etal-2022-revisiting}, which reduces the issue of false negatives.
These two datasets contain complex documents with an average of 19 entities and approximately 390 candidate pairs per text, for a total of 3,053 annotated Wikipedia articles in the training set. 
Beyond these benchmarks, other datasets have been proposed for specialized domains, such as HacRED \citep{cheng-etal-2021-hacred} for Chinese relation extraction and SciERC \citep{luan-etal-2018-multi} for scientific literature. However, to the best of our knowledge, DocRED and Re-DocRED remain the only widely adopted English benchmarks that provide evidence annotations, an essential feature for fair comparison with evidence-aware models such as DREEAM.

\paragraph{Few-Shot Document-level Relation Extraction}
Given the complexity of document-level relation extraction, the few-shot setting has received comparatively limited attention. A straightforward strategy involves training under data-constrained regimes, where only a small number of labeled samples are available. Moving beyond this paradigm, \citet{popovic-farber-2022-shot} formulate Few-Shot Document-level Relation Extraction (FSDRE) as a meta-learning problem and introduce the FREDo dataset, constructed upon DocRED and SciERC.
Models are evaluated on episodic tasks comprising support and query sets derived from DocRED in the original FREDo, and from Re-DocRED in its refined version, Re-FREDo \citep{meng-etal-2023-rapl}. SciERC is employed as an out-of-domain test set to evaluate cross-domain generalization. Approaches evaluated on these benchmarks, such as RAPL \citep{meng-etal-2023-rapl} and TPN \citep{zhang2024tpn}, typically employ prototype-based learning to derive robust class representations from only a handful of examples.

\section{Methodology}

\subsection{Document Level Relation Extraction}

Let a document be represented as a sequence of tokens. Within this document, there is a set of $m$ pre-identified entities $E = \{e_1, e_2, \dots, e_m\}$, where each entity $e_i$ is associated with one or more mentions in the text. 

The objective of DocRE is to identify valid relational triplets. Each triplet is of the form $(e_h, r, e_t)$, where $e_h, e_t \in E$ are the head and tail entities respectively ($h$ can be equal to $t$), and $r$ is the relation that holds between them. Based on this definition, DocRE is a multi-label classification task on every entity pair, where labels are the relation types.

\subsection{GLiDRE architecture}

GLiDRE architecture is inspired by the bi-encoder variant of GLiNER\footnote{\url{https://blog.knowledgator.com/meet-the-new-zero-shot-ner-architecture-30ffc2cb1ee0}}. The GLiDRE model, presented in Figure~\ref{fig:glidre_model}, uses two separate transformers, one for encoding the document and another for encoding relation labels, which offers several benefits: unlike the uni‑encoder version, the bi-encoder approach avoids concatenating labels and document tokens into a single input. Instead, we encode the document and labels separately. This eliminates the need for special separation tokens and overcomes encoder context length limits, permitting an effectively unlimited label set. Moreover, label embeddings can be precomputed, accelerating inference. However, the bi‑encoder architecture increases memory usage and lacks cross‑attention between labels. As a result, it may struggle to disambiguate semantically similar labels.

\paragraph{Label and word representations}

Label embeddings are obtained by mean-pooling over the token representations constituting label names. The computed label embeddings are passed through a two-layer feedforward network if the output dimension of the label encoder differs from the configured latent space dimension of GLiDRE.
For document tokens, we adopt the strategy used in standard NER models: for words split into subwords, we extract the representation of the first subword. 

\paragraph{Relation representations}

Given contextual embeddings $H \in \mathbb{R}^{L \times D}$ from the document encoder, where $L$ is the sequence length and $D$ is the hidden dimension of the document encoder, we construct a representation for each candidate relation defined by a pair of entities.

First, for each mention of an entity, the mention representation is computed by pooling the embeddings of its constituent words. 

If an entity has multiple mentions, its final representation $h_e$ is derived by taking the mean of all its mention representations. 

Finally, the representation $h_r$ for the relation between $e_h$ and $e_t$, is obtained by concatenating their respective entity representations and passing them through a two-layer feedforward network. 

\begin{equation} \label{eq:1}
h_r = \text{FFN}(h_{e_h} \otimes h_{e_t})
\end{equation}
where $\otimes$ denotes vector concatenation. 

\paragraph{Relation Scoring}

To determine whether a candidate entity pair $(e_h,e_t)$ instantiates a given relation type $t$, we measure the following matching score between the relation representation $h_r$ and the embedding of relation type $h_t$:
\begin{equation} \label{eq:2}
s(e_h, e_t, t) = \sigma(h_r^\top h_t)
\end{equation}
where $\sigma(x) = (1 + e^{-x})^{-1}$ is the sigmoid function. A relation $t$ is assigned to the entity pair $(e_h,e_t)$ if
\begin{equation} \label{eq:3}
s(e_h, e_t, t) > \tau
\end{equation}
where $\tau$ is a predefined decision threshold.

\subsection{Loss Function}
To address the issue of class imbalance within DocRE datasets, we use the Focal Loss function proposed by \citet{lin2017focal} during training. The Focal Loss is a dynamically scaled cross-entropy loss designed to down-weight the contribution of well-classified examples and concentrating the training on hard examples. It is defined as:
\begin{equation}  \label{eq:4}
    \text{FL}(p_t) = -\alpha_t (1 - p_t)^\gamma \log(p_t)
\end{equation}

\noindent where $p_t$ represents the model's estimated probability for the ground-truth class. The focusing parameter $\gamma$ adjusts the rate at which easy examples are down-weighted  and the weighting factor $\alpha_t$ balances the importance of positive and negative examples.

\subsection{Localized Context Pooling}

To create more refined representations, we also explore and adapt methods inspired by ATLOP \cite{zhou2021atlop} including Localized Context Pooling. Localized Context Pooling uses the attention scores from the encoder to focus on the most relevant parts of the document for a given entity pair.

Let $A \in \mathbb{R}^{L \times L}$ be the document-level attention matrix from the last encoder layer, averaged across all attention heads. For the head entity $e_h$ and tail entity $e_t$, we extract their corresponding attention vectors $A_h$ and $A_t$ by averaging the attention scores over their mention tokens. A joint attention vector $\alpha$ is computed via an element-wise product:
\begin{equation} \label{eq:5}
\alpha = A_h \odot A_t
\end{equation}
This $\alpha$ is then normalized and used to compute a weighted sum of the contextual embeddings, creating a localized context vector $c_{\text{loc}}$:
\begin{equation} \label{eq:6}
c_{\text{loc}} = \sum_{i=1}^{L} \frac{\alpha_i}{\sum_{j=1}^{L} \alpha_j} H_i
\end{equation}

Finally, the refined representations for the head $h'_{e_h}$ and tail $h'_{e_t}$ entities are created by concatenating their initial representations with the localized context vector and passing them through distinct two-layer feedforward network:

\begin{equation} \label{eq:7}
h'_{e_h} = \tanh(\text{FFN}_h(h_{e_h} \otimes c_{\text{loc}}))
\end{equation}
\begin{equation} \label{eq:8}
h'_{e_t} = \tanh(\text{FFN}_t(h_{e_t} \otimes c_{\text{loc}}))
\end{equation}

The final relation representation $h_r$ is then computed from $h'_{e_h}$ and $h'_{e_t}$ using Equation \ref{eq:1}.

\subsection{Pretraining data}

For a better generalization, GLiDRE benefits from pretraining on a large-scale generalist annotated corpus. We build such a corpus using a semi-automated annotation methodology inspired by \citet{zhou2023universalner}. Documents are first randomly sampled from FineWeb \cite{penedo2024fineweb}, a high-quality dataset built from filtered and deduplicated English Common Crawl archives.

We then prompt a LLM to generate annotations for both entities and the relations between them in a structured JSON format (we used the Mistral-Small-24B-Instruct-2501 model, the detailed prompt is provided in Appendix \ref{sec:appendix_pretraining}). The raw outputs are filtered to remove instances containing malformed JSON or documents that exceed a predefined length threshold.

The resulting dataset contains 136,404 documents with highly diverse labels, featuring 76,497 unique relation types. The most frequent relation types include types such as \texttt{IS\_LOCATED\_IN}, \texttt{WORKS\_FOR}, \texttt{PART\_OF} and \texttt{CONTAINS}. We use this dataset to pretrain our model.

\section{Experiments}

\subsection{Datasets}
We conduct our experiments on the English benchmark Re-DocRED \cite{tan-etal-2022-revisiting}, a human-revised version of the DocRED benchmark that addresses its high false-negative rate, logical inconsistencies and coreference errors by re‑annotating all 4,053 documents in the original training and evaluation splits. Re-DocRED retains the same relation schema but substantially increases the number of annotated relation triples per document. 

For evaluating performance in a formal few-shot meta-learning context, we additionally use the FReDo \citep{popovic-farber-2022-shot} and Re-FReDo \citep{meng-etal-2023-rapl} benchmarks. These datasets re-purpose documents from DocRED, Re-DocRED, and SciERC into an episodic format. The evaluation is structured into two tasks: an in-domain task with approximately 15k episodes derived from DocRED, and a more challenging cross-domain task with 3k episodes derived from SciERC. Each episode consists of a small support set (1 or 3 documents) and a query set. To ensure a rigorous few-shot evaluation, the relation types in the training, development, and test splits are disjoint.

\subsection{Evaluation Settings}

\paragraph{Standard supervised settings} We evaluate the performance of our model on Re-DocRED across 3 distinct experimental settings, with a particular focus on low-data regimes where its advantages are most pronounced. First, to assess data efficiency generalization, we finetune different models on randomly sampled subsets of the training data containing 1, 5, 10, 50, 100, 500, or 1000 documents. Second, in a fully supervised setting, we finetune on the entire training set and compare against strong baselines. Finally, we examine zero-shot performance by comparing against larger LLMs without any task-specific fine-tuning, highlighting the  competitiveness of the model in scenarios with little or no labeled data.

\paragraph{Episodic meta-learning settings} To evaluate our model in the FSDRE setting on the FReDo and Re-FReDo benchmarks, we adopted a two-stage fine-tuning protocol for a fair comparison with prototype-based methods. First, we perform an initial alignment phase by training the model for 500 steps on the full training set, mainly to align label representations (e.g., capitalization consistency). Subsequently, for each episode, we further finetune the model for 20 epochs on the provided 1-shot or 3-shot support set before evaluating its performance on the corresponding query set.

\paragraph{Evaluation Metrics} We use the standard metrics for the Re-DocRED dataset~\citep{tan-etal-2022-revisiting}: \textit{F1} is the micro-averaged F1 score on the relation triples, \textit{Ign\_F1} is the F1 score ignoring the triples in the test set that appear also in the training set.

\paragraph{Implementation Details}
All experiments are conducted on a single NVIDIA H100 GPU with a batch size of 16. For the pretraining phase, the model is trained for 50,000 steps. For finetuning, we train for 10,000 steps. We use two learning rates $1 \times 10^{-5}$ for the model encoders and $1 \times 10^{-4}$ for all other layers. For the low-resource and fully supervised experiments, we report the average and standard deviation over 5 runs with different seeds. The model checkpoint yielding the best performance on the development set and a decision threshold  of 0.5 are used for the evaluation on the test set. We use two English models: DebertaV3 Large \cite{he2021deberta} for document encoder and BGE-Large V1.5 \cite{bge_embedding} for label encoder, similar to the bi-encoder version of GLiNER. The final model comprises approximately 800 million parameters. The pretraining phase requires approximately 24 hours, while finetuning on the ReDocRED dataset takes 3.5 hours.
Additionally, evaluating our model on the Re-FReDo benchmark is computationally intensive due to the episodic protocol, which requires repeated fine-tuning and model weight resets. This process amounts to approximately 35 hours of computation for the in-domain task and 5 hours for the cross-domain task.

\subsection{Results}

We present in this section the main results of the evaluation of GLiDRE in several settings. An additional detailed analysis of some parameters of the model such as the effect of pretraining, pooling strategies and adaptive thresholding are presented in Appendix~\ref{app:analysis}. 

\subsubsection{Low-Resource Settings}
In low-resource scenarios, we evaluate the efficiency of our model by comparing it against two strong supervised baselines:

\textbf{DREEAM} \citep{ma-etal-2023-dreeam} represents state-of-the-art on the fully-supervised benchmark and relies on guiding the model attention with evidence information as supervisory signals in a memory‐efficient manner and can employs a self‐training strategy to learn evidence retrieval without explicit annotations.

\textbf{ATLOP} \citep{zhou2021atlop} formulates document-level relation extraction as a semantic‐segmentation task, introducing localized context pooling to capture entity‐focused local contexts and an adaptive thresholding loss to learn dynamic decision thresholds. It is more comparable to our approach since it does not requires additional evidence information.

This comparison is designed to determine the dataset size at which these conventional supervised methods can match the performance of our model and quantifying the advantages of our approach in data-scarce regimes.

\begin{table*}[ht!]
\centering
\scalebox{0.90}{\hspace{-0.5em}
\begin{tabular}{@{}lccccccc@{}}
\toprule
\textbf{Model} & \textbf{N = 1} & \textbf{N = 5} & \textbf{N = 10} & \textbf{N = 50} & \textbf{N = 100} & \textbf{N = 500} & \textbf{N = 1000} \\
\midrule
ATLOP & $4.32_{\pm 3.19}$ & $18.76_{\pm 4.93}$ & $29.48_{\pm 3.91}$ & $50.36_{\pm 1.18}$ & $57.17_{\pm 0.37}$ & $68.91_{\pm 0.43}$ & $71.70_{\pm 0.20}$ \\
DREEAM & $4.27_{\pm 3.50}$ & $16.02_{\pm 7.46}$ & $27.07_{\pm 5.81}$ & $52.05_{\pm 2.00}$ & $58.14_{\pm 0.86}$ & $\mathbf{69.32}_{\pm 0.31}$ & $71.67_{\pm 0.32}$ \\
\textbf{GLiDRE (Ours)} & $\mathbf{24.45}_{\pm 6.37}$ & $\mathbf{33.71}_{\pm 6.11}$ & $\mathbf{41.73}_{\pm 2.94}$ & $\mathbf{54.54}_{\pm 1.39}$ & $\mathbf{60.04}_{\pm 0.50}$ & $69.26_{\pm 0.37}$ & $\mathbf{72.09}_{\pm 0.31}$ \\
\bottomrule
\end{tabular}}
\caption{\textbf{Low-resource supervised setting}: micro-averaged F1 results for GLiDRE, ATLOP and DREEAM across various training set sizes ($N$). All models are trained on the same 5 subsets for each value of $N$. Best results per column are shown in \textbf{bold}.}
\label{tab:few_shot_results}
\end{table*}

The results presented in Table \ref{tab:few_shot_results} clearly demonstrate the superiority of GLiDRE in data-scarce environments. In the extremely low-data regime (N $\le$ 100), our model establishes a lead over both ATLOP and the state-of-the-art model, DREEAM. For instance, with only 10 training samples, GLiDRE achieves an F1 score of $41.73$, surpassing DREEAM by a margin of over 14 F1 points. This significant advantage persists up to N=100 samples, the performance gap narrowing only when the amount of training data increases to 500 and 1000 samples.

\subsubsection{Few-Shot Document-level Relation Extraction}
\begin{table*}[ht!]
\centering
\resizebox{\linewidth}{!}{
\begin{tabular}{lcccccccc}
\toprule
\multirow{3}{*}{\textbf{Model}}    & \multicolumn{4}{c}{\textbf{FREDo}}                         & \multicolumn{4}{c}{\textbf{ReFREDo}}                      \\  \cmidrule(r){2-5} \cmidrule(r){6-9}
                           & \multicolumn{2}{c}{\textbf{In-Domain}} & \multicolumn{2}{c}{\textbf{Cross-Domain}} & \multicolumn{2}{c}{\textbf{In-Domain}} & \multicolumn{2}{c}{\textbf{Cross-Domain}} \\  \cmidrule(r){2-3} \cmidrule(r){4-5}  \cmidrule(r){6-7} \cmidrule(r){8-9}
& 1-Doc $F_1$ & 3-Doc $F_1$ & 1-Doc $F_1$ & 3-Doc $F_1$ & 1-Doc $F_1$ & 3-Doc $F_1$ & 1-Doc $F_1$ & 3-Doc $F_1$  \\
\midrule
DL-Base & 0.60 & 0.89 & 1.76 & 1.98 & 1.38 & 1.84 & 1.76 & 1.98  \\
DL-MNAV & 7.05 & 8.42 & 0.84 & 0.48 & 12.97 & 12.43  & 1.12 & 2.28 \\
$\textnormal{DL-MNAV}_{SIE}$ & 7.06 & 6.77 & 1.77 & 2.51 & 13.37 & 12.00 & 1.39 & 2.92 \\
$\textnormal{DL-MNAV}_{SIE+SBN}$ & 1.71 & 2.79 & 2.85 & 3.72 & 4.59 & 5.43 & 2.84 & 3.86 \\
RAPL & 8.75 & 10.67 & 3.33 & 5.35 & 15.20 & 16.35 & 3.51 & 5.48  \\
TPN & 9.12 & 8.64 & 3.98 & 4.48 & 15.54 & 15.73 & 4.72 & 5.02  \\
\textbf{GLiDRE (Ours)} & \textbf{13.00} & \textbf{15.71} & \textbf{10.97} & \textbf{11.00} & \textbf{26.36} & \textbf{28.77} & \textbf{10.60} & \textbf{10.32}  \\
\bottomrule
\end{tabular}
}
\caption{\textbf{Few-shot episodic setting}: results on FREDo and ReFREDo benchmarks. Reported scores are macro-averaged across relation types and obtained from \citet{meng-etal-2023-rapl} and \citet{zhang2024tpn}. Best results per column are shown in \textbf{bold}.}
\label{tab:fredo_results}
\end{table*}
For the FSDRE evaluation, we follow the episodic protocols defined by FREDo and Re-FREDo. We compare GLiDRE to three methods specifically developed for episodic few-shot DocRE:

\textbf{DL-MNAV} \citep{popovic-farber-2022-shot} adapts sentence-level few-shot relation extraction MNAV~\cite{sabo-etal-2021-revisiting} to documents by pooling mention representations and explicitly modeling NOTA (\textit{none-of-the-above}) via learned NOTA vectors, adaptive-threshold loss, and support-based NOTA sampling at inference.  

\textbf{RAPL} \citep{meng-etal-2023-rapl}  refines prototypes with instance-level aggregation and relation-weighted contrastive learning, and constructs task-specific NOTA prototypes. 

\textbf{TPN} \citep{zhang2024tpn} improves cross-domain transfer with a hybrid encoder, transferable NOTA prototype learning, and a calibration module to mitigate NOTA bias.


As shown in Table \ref{tab:fredo_results}, GLiDRE establishes a new state-of-the-art on both benchmarks by a substantial margin across all settings. The advantage of our model is particularly pronounced in the challenging cross-domain setting, where the task is to generalize to relations from the SciERC dataset, highlighting the effectiveness of our synthetic data pretraining strategy for FSDRE and domain transfer.

\begin{table*}[ht!]
\centering
\caption{\textbf{Fully supervised setting}: results on the Re-DocRED dataset. We compare GLiDRE with strong DocRE models, LLM-based methods and GLiREL. F1 scores are reported on both the development and test sets. Results are taken from the original papers, except for DocRE models, which are reported following \citet{gao-etal-2024-ttm}.}
\label{tab:fully_supervised_results}
\begin{tabular}{lcccc}
\toprule
\textbf{Model} & \textbf{Dev F1} & \textbf{Dev Ign F1} & \textbf{Test F1 } & \textbf{Test Ign F1} \\
\midrule
\multicolumn{5}{l}{\textit{DocRE Models}} \\
ATLOP \citep{zhou2021atlop} & $76.15_{\pm 0.23}$ & $75.88_{\pm 0.23}$ & $77.81_{\pm 0.71}$ & $76.13_{\pm 0.28}$\\
KD-DocRE \citep{tan-etal-2022-document} & $77.88_{\pm 0.42}$ & $77.12_{\pm 0.49}$ & $78.28_{\pm 0.72}$ & $77.60_{\pm 0.25}$ \\
TTM-RE \citep{gao-etal-2024-ttm} & $78.13_{\pm 0.12}$ & $78.05_{\pm 0.17}$ & $79.95_{\pm 0.13}$ & $78.20_{\pm 0.34}$ \\ 
DREEAM \citep{ma-etal-2023-dreeam} & $79.42_{\pm 0.18}$ & $78.36_{\pm 0.19}$ & $80.20_{\pm 0.45}$ & $78.56_{\pm 0.39}$ \\
\midrule
\multicolumn{5}{l}{\textit{LLM-based \citep{li2024llm}}} \\
LMRC LLaMA2-13B-Chat & - & - & $74.63$ & $74.08$ \\
\midrule
\multicolumn{5}{l}{\textit{GLiNER-inspired model}} \\
GLiREL \citep{boylan-etal-2025-glirel} & - & - & $54.13$ & $53.24$ \\
\midrule
\textbf{GLiDRE (Ours)} & $77.76_{\pm 0.35}$ & $76.70_{\pm 0.37}$ & $77.83_{\pm 0.23}$ & $76.80_{\pm 0.22}$\\
\bottomrule
\end{tabular}
\end{table*}

\subsubsection{Fully supervised setting} \label{sec:fully_supervised_results}
Additionally, we evaluate the scalability of our proposed model, \textbf{GLiDRE}, by comparing it against several strong baselines on the Re-DocRED dataset under a fully supervised protocol. Specifically, in addition to \textbf{DREEAM} and \textbf{ATLOP}, we also include the following baselines:

\textbf{KD-DocRE} \citep{tan-etal-2022-document} leverages an axial attention module to model entity-pairs interdependency across sentences, applies an adaptive focal loss to mitigate class imbalance and uses knowledge distillation to incorporate distantly supervised data.

\textbf{TTM-RE} \citep{gao-etal-2024-ttm} introduces a Token Turing Machine memory module that augments document representations with external memory tokens, paired with a noise‐robust loss tailored for positive–unlabeled setting.

Additionally, we benchmark against \textbf{GLiREL} \citep{boylan-etal-2025-glirel}, a sentence-level relation extraction adaptation of GLiNER architecture. It is important to note that GLiREL was originally conceived for sentence-level relation extraction and operates with a shorter context length. For document-level predictions, it relies on aggregating relations based on gold-standard coreference information.

We also report recent methods that leverage LLMs, comparing with the framework introduced by \citet{li2024llm}, which involves fine-tuning Llama models \citep{touvron2023llama} using Low-Rank Adaptation \citep{hu2022lora} and enhancing the model performance by employing a classifier to exhibit potential relations and guide the finetuned LLM. We only report their best result on this dataset.

The results presented in Table~\ref{tab:fully_supervised_results}, demonstrate that GLiDRE achieves competitive performance on the Re-DocRED benchmark. Notably, this performance is attained despite the model not being specifically designed for large-scale datasets, highlighting its versatility and suitability across diverse scenarios. Our model attains a test F1 score of \textbf{$77.83$}. This places GLiDRE in close competition with established DocRE models like ATLOP and KD-DocRE. While it does not surpass the current state-of-the-art methods like DREEAM and TTM-RE, it is important to note that these models incorporate additional mechanisms, such as evidence information for DREEAM and the Turing Token Machine for TTM-RE that is designed to scale better than our approach with large datasets.

Critically, GLiDRE outperforms LLM-based methods. It surpasses the best-performing LMRC approach by over 3 F1 points, despite being a significantly smaller and more computationally efficient model. This highlights the strength of specialized encoder architectures for this task compared to fine-tuning general-purpose LLMs.

Furthermore, the comparison with GLiREL highlights the importance of our architectural adaptations for the document-level context. By designing representations specifically for relation extraction at the document level, GLiDRE achieves an improvement of over 23 F1 points.

\subsubsection{Zero-Shot Setting}
\begin{table*}[!ht]
\centering
\caption{\textbf{Zero-shot setting}: results on the Re-DocRED dev and test sets. Results for GPT-3.5 Turbo are from \citet{xue-etal-2024-autore}. All other LLMs are instruction-tuned versions.}
\label{tab:zeroshot_comparison}
\begin{tabular}{lcccc}
\toprule
\textbf{Model} & \textbf{Dev F1} & \textbf{Dev Ign F1} & \textbf{Test F1} & \textbf{Test Ign F1} \\
\midrule
\multicolumn{5}{l}{\textit{Large Language Model Baselines}} \\
Qwen 2.5 72B & 18.24 & 18.09 & 18.00 & 17.86 \\
Llama 3.3 70B & 15.67 & 15.57 & 15.81 & 15.71 \\
Mistral-Large 123B & \textbf{18.49} & \textbf{18.33} & \textbf{18.61} & \textbf{18.50} \\
Mistral-Small 24B & 14.39 & 14.27 & 14.38 & 14.27 \\
GPT-3.5 Turbo & - & - & 6.68 & - \\
\midrule
\multicolumn{5}{l}{\textit{Proposed Model}} \\
\textbf{GLiDRE} & 16.72 & 16.37 & 17.32 & 16.41 \\
\bottomrule
\end{tabular}
\end{table*}
The zero-shot setting for Document-Level Relation Extraction remains a largely underexplored research area. Given the scarcity of existing baselines, we establish a benchmark by evaluating the performance of much larger open-weight Large Language Models. We compare against recent high-performing models, specifically employing the instruction-tuned versions of Qwen 2.5 72B \citep{qwen2.5}, Mistral Large 123B\footnote{\url{https://huggingface.co/mistralai/Mistral-Large-Instruct-2411}}, Llama 3.3 70B\footnote{\url{https://huggingface.co/meta-llama/Llama-3.3-70B-Instruct}} and the model we used for pretraining data generation, Mistral-Small 24B. Our prompting methodology is inspired by recent work in zero-shot Information Extraction \citep{yuan2023zero}. We annotate entity mentions directly within the input text using special tags and, in order to ensure structured and reliable outputs, we constrain the generation process to a strict \texttt{(head, relation, tail)} triplet format. This constrained decoding is implemented using regular expressions within the vLLM inference framework \citep{kwon2023efficient}, which facilitates parsing and mitigates the risk of hallucination. The detailed prompt template used for this task is provided in Appendix \ref{sec:appendix_prompt_zero_shot}. We use a fixed temperature of 0.

The results of our zero-shot evaluation, detailed in Table \ref{tab:zeroshot_comparison}, highlight the efficiency and effectiveness of GLiDRE. We establish a strong benchmark by comparing our model against several state-of-the-art LLMs.

Our primary finding is that GLiDRE, a model with only a few hundred million parameters, achieves performance that is competitive with LLMs orders of magnitude larger. With a test F1 score of $17.32$, GLiDRE outperforms Llama 3.3 70B and Mistral-Small 24B (which was used to generate the pretraining data) demonstrating the effectiveness to train on the pretraining corpus, as opposed to directly leveraging Mistral-Small 24B to annotate DocRED. This demonstrates that for specialized tasks like DocRE, a focused bi-encoder architecture can rival the zero-shot reasoning capabilities of massive general-purpose models. Furthermore, the results show a dramatic improvement when compared to the score of GPT-3.5 Turbo reported by \citet{xue-etal-2024-autore}, showing the rapid recent progress of LLMs in zero-shot Information Extraction tasks.
    
Beyond performance, GLiDRE offers significant practical advantages. The inference process on the Re-DocRED test set (500 documents) with GLiDRE (0.8B parameters) requires less than 10GB of VRAM and completes in approximately 100 seconds on a single NVIDIA A100-80GB GPU. In contrast, inference with Mistral-Large (123B parameters) necessitates a distributed setup of at least 4x A100-80GB GPUs to accommodate its memory footprint of over 300GB. This process takes around 600 seconds and a total cost of approximately 2400 GPU-seconds. This breakdown demonstrates that GLiDRE is not only smaller but over 20 times more cost-effective in terms of total compute required for inference. Moreover, using LLMs for specific tasks often require complex engineering, including sophisticated prompting and constrained decoding, to produce structured, parsable outputs and mitigate hallucinations, while our model natively produces structured predictions.

The performance of GLiDRE shows some variability, which may come from a misalignment between their general-purpose pretraining corpora and the specific domain and relations of the Re-DocRED dataset. Adapting its pretraining by generating synthetic data with relation types aligned to the target dataset, as was done with GLiNER for Personally Identifiable Informations\footnote{\url{https://huggingface.co/urchade/gliner_multi_pii-v1}}, could improve results. However, this would likely reduce its ability to generalize to other DocRE datasets.

\section{Conclusion}
We present GLiDRE, a novel lightweight bi‑encoder model for document‑level relation extraction that reconceptualizes Document-level Relation Extraction as a direct representation matching problem. We show, through extensive experiments on the Re‑DocRED, FREDo and Re-FREDo benchmarks, that GLiDRE achieves state‑of‑the‑art few‑shot performance and matches or surpasses much larger LLMs in zero‑shot settings, all while operating with a fraction of their computational footprint.

GLiDRE not only rivals fully supervised baselines under low‑resource regimes but also delivers highly structured predictions without complex prompting or constrained decoding. Its efficiency makes it a practical choice for real‑world IE applications.

Future work will explore synthetic relation generation to further close remaining performance gaps and investigate dynamic thresholding methods tailored to the bi‑encoder setup to enhance robustness across diverse relation schemas.

\section*{Limitations}

Despite its efficiency and strong few‑shot performance, GLiDRE faces several inherent limitations.

The number of candidate relation pairs grows quadratically with the number of entities in a document, which can dramatically increase memory consumption for texts with high entity density and even lead to out‑of‑memory errors. 

Although the bi‑encoder design can in principle accommodate longer contexts, it remains bound by the maximum sequence length of the underlying document encoder (e.g. 512 tokens for DeBERTa). Other approaches we compare against, such as DREEAM, ATLOP, KD-DocRE, and TTM-RE, have the same limitation. Documents that exceed this limit must be truncated or segmented into chunks, potentially disrupting long‑distance dependencies and harming performance.

The independent scoring of each entity‐relation pair in GLiDRE overlooks inter‑label interactions. As a result, the model can struggle to distinguish semantically similar relation types in the absence of a joint classification mechanism. 

Our evaluation follows standard DocRE protocols by assuming gold entities and coreference chains are provided. Since GLiDRE does not perform named entity recognition or coreference resolution, its effectiveness in a fully end‑to‑end pipeline would depend on the accuracy of upstream modules, which may propagate errors and degrade overall performance. This limitation is shared by other existing baselines and is particularly pronounced for DREEAM, which additionally relies on evidence annotations.

The pretraining of GLiDRE relies on the synthetic annotation of texts with a LLM, which is a source of risk for the propagation of the inherent biases of the LLM. This risk is mitigated by the use of clear annotation guidelines and could be further limited by combining multiple prompts and models and having human reviewers audit a subset of the annotations for bias and correctness.

\section*{Acknowledgments}

This publication was made possible by the use of the FactoryIA supercomputer, financially supported by the Ile-De-France Regional Council. It also benefited from the support of the DataFIX project, financed by the French government under the France 2030 Programme and operated by Bpifrance.

\bibliography{anthology, custom}

\begin{thebibliography}{44}
\providecommand{\natexlab}[1]{#1}

\bibitem[{Boylan et~al.(2025)Boylan, Hokamp, and Ghalandari}]{boylan-etal-2025-glirel}
Jack Boylan, Chris Hokamp, and Demian~Gholipour Ghalandari. 2025.
\newblock \href {https://doi.org/10.18653/v1/2025.naacl-long.418} {{GL}i{REL} - generalist model for zero-shot relation extraction}.
\newblock In \emph{Proceedings of the 2025 Conference of the Nations of the Americas Chapter of the Association for Computational Linguistics: Human Language Technologies (Volume 1: Long Papers)}, pages 8230--8245, Albuquerque, New Mexico. Association for Computational Linguistics.

\bibitem[{Chen and Li(2021)}]{chen-li-2021-zs}
Chih-Yao Chen and Cheng-Te Li. 2021.
\newblock \href {https://doi.org/10.18653/v1/2021.naacl-main.272} {{ZS}-{BERT}: Towards zero-shot relation extraction with attribute representation learning}.
\newblock In \emph{Proceedings of the 2021 Conference of the North American Chapter of the Association for Computational Linguistics: Human Language Technologies}, pages 3470--3479, Online. Association for Computational Linguistics.

\bibitem[{Cheng et~al.(2021)Cheng, Liu, Qu, Zhao, Liang, Wang, Huai, Yuan, and Xiao}]{cheng-etal-2021-hacred}
Qiao Cheng, Juntao Liu, Xiaoye Qu, Jin Zhao, Jiaqing Liang, Zhefeng Wang, Baoxing Huai, Nicholas~Jing Yuan, and Yanghua Xiao. 2021.
\newblock \href {https://doi.org/10.18653/v1/2021.findings-acl.249} {{H}ac{RED}: A large-scale relation extraction dataset toward hard cases in practical applications}.
\newblock In \emph{Findings of the Association for Computational Linguistics: ACL-IJCNLP 2021}, pages 2819--2831, Online. Association for Computational Linguistics.

\bibitem[{Chia et~al.(2022)Chia, Bing, Poria, and Si}]{chia-etal-2022-relationprompt}
Yew~Ken Chia, Lidong Bing, Soujanya Poria, and Luo Si. 2022.
\newblock \href {https://doi.org/10.18653/v1/2022.findings-acl.5} {{R}elation{P}rompt: Leveraging prompts to generate synthetic data for zero-shot relation triplet extraction}.
\newblock In \emph{Findings of the Association for Computational Linguistics: ACL 2022}, pages 45--57, Dublin, Ireland. Association for Computational Linguistics.

\bibitem[{Gao et~al.(2024)Gao, Wang, and Sun}]{gao-etal-2024-ttm}
Chufan Gao, Xuan Wang, and Jimeng Sun. 2024.
\newblock \href {https://doi.org/10.18653/v1/2024.acl-long.26} {{TTM}-{RE}: Memory-augmented document-level relation extraction}.
\newblock In \emph{Proceedings of the 62nd Annual Meeting of the Association for Computational Linguistics (Volume 1: Long Papers)}, pages 443--458, Bangkok, Thailand. Association for Computational Linguistics.

\bibitem[{Gao et~al.(2019)Gao, Han, Zhu, Liu, Li, Sun, and Zhou}]{gao-etal-2019-fewrel}
Tianyu Gao, Xu~Han, Hao Zhu, Zhiyuan Liu, Peng Li, Maosong Sun, and Jie Zhou. 2019.
\newblock \href {https://doi.org/10.18653/v1/D19-1649} {{F}ew{R}el 2.0: Towards more challenging few-shot relation classification}.
\newblock In \emph{Proceedings of the 2019 Conference on Empirical Methods in Natural Language Processing and the 9th International Joint Conference on Natural Language Processing (EMNLP-IJCNLP)}, pages 6250--6255, Hong Kong, China. Association for Computational Linguistics.

\bibitem[{Han et~al.(2018)Han, Zhu, Yu, Wang, Yao, Liu, and Sun}]{han-etal-2018-fewrel}
Xu~Han, Hao Zhu, Pengfei Yu, Ziyun Wang, Yuan Yao, Zhiyuan Liu, and Maosong Sun. 2018.
\newblock \href {https://doi.org/10.18653/v1/D18-1514} {{F}ew{R}el: A large-scale supervised few-shot relation classification dataset with state-of-the-art evaluation}.
\newblock In \emph{Proceedings of the 2018 Conference on Empirical Methods in Natural Language Processing}, pages 4803--4809, Brussels, Belgium. Association for Computational Linguistics.

\bibitem[{He et~al.(2021)He, Liu, Gao, and Chen}]{he2021deberta}
Pengcheng He, Xiaodong Liu, Jianfeng Gao, and Weizhu Chen. 2021.
\newblock \href {https://openreview.net/forum?id=XPZIaotutsD} {Deberta: Decoding-enhanced bert with disentangled attention}.
\newblock In \emph{International Conference on Learning Representations}.

\bibitem[{Hendrickx et~al.(2010)Hendrickx, Kim, Kozareva, Nakov, {\'O}~S{\'e}aghdha, Pad{\'o}, Pennacchiotti, Romano, and Szpakowicz}]{hendrickx-etal-2010-semeval}
Iris Hendrickx, Su~Nam Kim, Zornitsa Kozareva, Preslav Nakov, Diarmuid {\'O}~S{\'e}aghdha, Sebastian Pad{\'o}, Marco Pennacchiotti, Lorenza Romano, and Stan Szpakowicz. 2010.
\newblock \href {https://aclanthology.org/S10-1006/} {{S}em{E}val-2010 task 8: Multi-way classification of semantic relations between pairs of nominals}.
\newblock In \emph{Proceedings of the 5th International Workshop on Semantic Evaluation}, pages 33--38, Uppsala, Sweden. Association for Computational Linguistics.

\bibitem[{Hu et~al.(2022)Hu, Shen, Wallis, Allen-Zhu, Li, Wang, Wang, Chen et~al.}]{hu2022lora}
Edward~J Hu, Yelong Shen, Phillip Wallis, Zeyuan Allen-Zhu, Yuanzhi Li, Shean Wang, Lu~Wang, Weizhu Chen, and 1 others. 2022.
\newblock Lora: Low-rank adaptation of large language models.
\newblock \emph{ICLR}, 1(2):3.

\bibitem[{Jia et~al.(2019)Jia, Wong, and Poon}]{jia-etal-2019-document}
Robin Jia, Cliff Wong, and Hoifung Poon. 2019.
\newblock \href {https://doi.org/10.18653/v1/N19-1370} {Document-level n-ary relation extraction with multiscale representation learning}.
\newblock In \emph{Proceedings of the 2019 Conference of the North {A}merican Chapter of the Association for Computational Linguistics: Human Language Technologies, Volume 1 (Long and Short Papers)}, pages 3693--3704, Minneapolis, Minnesota. Association for Computational Linguistics.

\bibitem[{Kwon et~al.(2023)Kwon, Li, Zhuang, Sheng, Zheng, Yu, Gonzalez, Zhang, and Stoica}]{kwon2023efficient}
Woosuk Kwon, Zhuohan Li, Siyuan Zhuang, Ying Sheng, Lianmin Zheng, Cody~Hao Yu, Joseph Gonzalez, Hao Zhang, and Ion Stoica. 2023.
\newblock Efficient memory management for large language model serving with pagedattention.
\newblock In \emph{Proceedings of the 29th symposium on operating systems principles}, pages 611--626.

\bibitem[{Lan et~al.(2023)Lan, Li, Zhang, Zhao, and Zhao}]{MC-BERT-Lan-et-al}
Yuquan Lan, Dongxu Li, Yunqi Zhang, Hui Zhao, and Gang Zhao. 2023.
\newblock \href {https://doi.org/10.1109/IJCNN54540.2023.10191459} {Modeling zero-shot relation classification as a multiple-choice problem}.
\newblock In \emph{2023 International Joint Conference on Neural Networks (IJCNN)}, pages 1--8.

\bibitem[{Li et~al.(2024{\natexlab{a}})Li, Wang, Liu, Guo, Ji, Shang, and Xu}]{ijcai2024p702}
Guozheng Li, Peng Wang, Jiajun Liu, Yikai Guo, Ke~Ji, Ziyu Shang, and Zijie Xu. 2024{\natexlab{a}}.
\newblock \href {https://doi.org/10.24963/ijcai.2024/702} {Meta in-context learning makes large language models better zero and few-shot relation extractors}.
\newblock In \emph{Proceedings of the Thirty-Third International Joint Conference on Artificial Intelligence, {IJCAI-24}}, pages 6350--6358. International Joint Conferences on Artificial Intelligence Organization.
\newblock Main Track.

\bibitem[{Li et~al.(2023)Li, Jia, and Zheng}]{li-etal-2023-semi}
Junpeng Li, Zixia Jia, and Zilong Zheng. 2023.
\newblock \href {https://doi.org/10.18653/v1/2023.emnlp-main.334} {Semi-automatic data enhancement for document-level relation extraction with distant supervision from large language models}.
\newblock In \emph{Proceedings of the 2023 Conference on Empirical Methods in Natural Language Processing}, pages 5495--5505, Singapore. Association for Computational Linguistics.

\bibitem[{Li et~al.(2024{\natexlab{b}})Li, Chen, Long, and Zhang}]{li2024llm}
Xingzuo Li, Kehai Chen, Yunfei Long, and Min Zhang. 2024{\natexlab{b}}.
\newblock Llm with relation classifier for document-level relation extraction.
\newblock \emph{arXiv preprint arXiv:2408.13889}.

\bibitem[{Lin et~al.(2017)Lin, Goyal, Girshick, He, and Doll{\'a}r}]{lin2017focal}
Tsung-Yi Lin, Priya Goyal, Ross Girshick, Kaiming He, and Piotr Doll{\'a}r. 2017.
\newblock Focal loss for dense object detection.
\newblock In \emph{Proceedings of the IEEE international conference on computer vision}, pages 2980--2988.

\bibitem[{Liu et~al.(2023)Liu, Zhu, Zhang, Feng, Chen, and Li}]{liu-etal-2023-document-level}
Yichun Liu, Zizhong Zhu, Xiaowang Zhang, Zhiyong Feng, Daoqi Chen, and Yaxin Li. 2023.
\newblock \href {https://doi.org/10.18653/v1/2023.emnlp-main.138} {Document-level relationship extraction by bidirectional constraints of beta rules}.
\newblock In \emph{Proceedings of the 2023 Conference on Empirical Methods in Natural Language Processing}, pages 2256--2266, Singapore. Association for Computational Linguistics.

\bibitem[{Luan et~al.(2018)Luan, He, Ostendorf, and Hajishirzi}]{luan-etal-2018-multi}
Yi~Luan, Luheng He, Mari Ostendorf, and Hannaneh Hajishirzi. 2018.
\newblock \href {https://doi.org/10.18653/v1/D18-1360} {Multi-task identification of entities, relations, and coreference for scientific knowledge graph construction}.
\newblock In \emph{Proceedings of the 2018 Conference on Empirical Methods in Natural Language Processing}, pages 3219--3232, Brussels, Belgium. Association for Computational Linguistics.

\bibitem[{Lv et~al.(2023)Lv, Liu, Dai, Liu, Yang, Luo, and Yu}]{lv-etal-2023-dsp}
Bo~Lv, Xin Liu, Shaojie Dai, Nayu Liu, Fan Yang, Ping Luo, and Yue Yu. 2023.
\newblock \href {https://doi.org/10.18653/v1/2023.findings-acl.339} {{DSP}: Discriminative soft prompts for zero-shot entity and relation extraction}.
\newblock In \emph{Findings of the Association for Computational Linguistics: ACL 2023}, pages 5491--5505, Toronto, Canada. Association for Computational Linguistics.

\bibitem[{Ma et~al.(2023)Ma, Wang, and Okazaki}]{ma-etal-2023-dreeam}
Youmi Ma, An~Wang, and Naoaki Okazaki. 2023.
\newblock \href {https://doi.org/10.18653/v1/2023.eacl-main.145} {{DREEAM}: Guiding attention with evidence for improving document-level relation extraction}.
\newblock In \emph{Proceedings of the 17th Conference of the European Chapter of the Association for Computational Linguistics}, pages 1971--1983, Dubrovnik, Croatia. Association for Computational Linguistics.

\bibitem[{Meng et~al.(2023)Meng, Hu, Liu, Li, Ma, Yang, and Wen}]{meng-etal-2023-rapl}
Shiao Meng, Xuming Hu, Aiwei Liu, Shuang Li, Fukun Ma, Yawen Yang, and Lijie Wen. 2023.
\newblock \href {https://doi.org/10.18653/v1/2023.emnlp-main.316} {{RAPL}: A relation-aware prototype learning approach for few-shot document-level relation extraction}.
\newblock In \emph{Proceedings of the 2023 Conference on Empirical Methods in Natural Language Processing}, pages 5208--5226, Singapore. Association for Computational Linguistics.

\bibitem[{M{\"o}ller and Usbeck(2024)}]{moller2024incorporating}
Cedric M{\"o}ller and Ricardo Usbeck. 2024.
\newblock Incorporating type information into zero-shot relation extraction.
\newblock In \emph{Proceedings of the Third International Workshop on Knowledge Graph Generation from Text (TEXT2KG 2024)}, pages 26--30.

\bibitem[{Penedo et~al.(2024)Penedo, Kydl{\'\i}{\v{c}}ek, Lozhkov, Mitchell, Raffel, Von~Werra, Wolf et~al.}]{penedo2024fineweb}
Guilherme Penedo, Hynek Kydl{\'\i}{\v{c}}ek, Anton Lozhkov, Margaret Mitchell, Colin~A Raffel, Leandro Von~Werra, Thomas Wolf, and 1 others. 2024.
\newblock The fineweb datasets: Decanting the web for the finest text data at scale.
\newblock \emph{Advances in Neural Information Processing Systems}, 37:30811--30849.

\bibitem[{Popovic and F{\"a}rber(2022)}]{popovic-farber-2022-shot}
Nicholas Popovic and Michael F{\"a}rber. 2022.
\newblock \href {https://doi.org/10.18653/v1/2022.naacl-main.421} {Few-shot document-level relation extraction}.
\newblock In \emph{Proceedings of the 2022 Conference of the North American Chapter of the Association for Computational Linguistics: Human Language Technologies}, pages 5733--5746, Seattle, United States. Association for Computational Linguistics.

\bibitem[{Sabo et~al.(2021)Sabo, Elazar, Goldberg, and Dagan}]{sabo-etal-2021-revisiting}
Ofer Sabo, Yanai Elazar, Yoav Goldberg, and Ido Dagan. 2021.
\newblock \href {https://doi.org/10.1162/tacl_a_00392} {Revisiting few-shot relation classification: Evaluation data and classification schemes}.
\newblock \emph{Transactions of the Association for Computational Linguistics}, 9:691--706.

\bibitem[{Sainz et~al.(2024)Sainz, Garc{\'\i}a-Ferrero, Agerri, de~Lacalle, Rigau, and Agirre}]{sainz2024gollie}
Oscar Sainz, Iker Garc{\'\i}a-Ferrero, Rodrigo Agerri, Oier~Lopez de~Lacalle, German Rigau, and Eneko Agirre. 2024.
\newblock \href {https://openreview.net/forum?id=Y3wpuxd7u9} {Go{LLIE}: Annotation guidelines improve zero-shot information-extraction}.
\newblock In \emph{The Twelfth International Conference on Learning Representations}.

\bibitem[{Tan et~al.(2022{\natexlab{a}})Tan, He, Bing, and Ng}]{tan-etal-2022-document}
Qingyu Tan, Ruidan He, Lidong Bing, and Hwee~Tou Ng. 2022{\natexlab{a}}.
\newblock \href {https://doi.org/10.18653/v1/2022.findings-acl.132} {Document-level relation extraction with adaptive focal loss and knowledge distillation}.
\newblock In \emph{Findings of the Association for Computational Linguistics: ACL 2022}, pages 1672--1681, Dublin, Ireland. Association for Computational Linguistics.

\bibitem[{Tan et~al.(2022{\natexlab{b}})Tan, Xu, Bing, Ng, and Aljunied}]{tan-etal-2022-revisiting}
Qingyu Tan, Lu~Xu, Lidong Bing, Hwee~Tou Ng, and Sharifah~Mahani Aljunied. 2022{\natexlab{b}}.
\newblock \href {https://doi.org/10.18653/v1/2022.emnlp-main.580} {Revisiting {D}oc{RED} - addressing the false negative problem in relation extraction}.
\newblock In \emph{Proceedings of the 2022 Conference on Empirical Methods in Natural Language Processing}, pages 8472--8487, Abu Dhabi, United Arab Emirates. Association for Computational Linguistics.

\bibitem[{Team(2024)}]{qwen2.5}
Qwen Team. 2024.
\newblock \href {https://qwenlm.github.io/blog/qwen2.5/} {Qwen2.5: A party of foundation models}.

\bibitem[{Touvron et~al.(2023)Touvron, Lavril, Izacard, Martinet, Lachaux, Lacroix, Rozi{\`e}re, Goyal, Hambro, Azhar et~al.}]{touvron2023llama}
Hugo Touvron, Thibaut Lavril, Gautier Izacard, Xavier Martinet, Marie-Anne Lachaux, Timoth{\'e}e Lacroix, Baptiste Rozi{\`e}re, Naman Goyal, Eric Hambro, Faisal Azhar, and 1 others. 2023.
\newblock Llama: Open and efficient foundation language models.
\newblock \emph{arXiv preprint arXiv:2302.13971}.

\bibitem[{Tran et~al.(2023)Tran, Ouchi, Shindo, Matsumoto, and Watanabe}]{tran2023enhancing}
Van-Hien Tran, Hiroki Ouchi, Hiroyuki Shindo, Yuji Matsumoto, and Taro Watanabe. 2023.
\newblock \href {https://doi.org/10.5715/jnlp.30.304} {Enhancing semantic correlation between instances and relations for zero-shot relation extraction}.
\newblock \emph{Journal of Natural Language Processing}, 30(2):304--329.

\bibitem[{Wang et~al.(2023)Wang, Zhou, Zu, Xia, Chen, Zhang, Zheng, Ye, Zhang, Gui et~al.}]{wang2023instructuie}
Xiao Wang, Weikang Zhou, Can Zu, Han Xia, Tianze Chen, Yuansen Zhang, Rui Zheng, Junjie Ye, Qi~Zhang, Tao Gui, and 1 others. 2023.
\newblock Instructuie: Multi-task instruction tuning for unified information extraction.
\newblock \emph{arXiv preprint arXiv:2304.08085}.

\bibitem[{Wei et~al.(2023)Wei, Cui, Cheng, Wang, Zhang, Huang, Xie, Xu, Chen, Zhang et~al.}]{wei2023zero}
Xiang Wei, Xingyu Cui, Ning Cheng, Xiaobin Wang, Xin Zhang, Shen Huang, Pengjun Xie, Jinan Xu, Yufeng Chen, Meishan Zhang, and 1 others. 2023.
\newblock Zero-shot information extraction via chatting with chatgpt.
\newblock \emph{arXiv preprint arXiv:2302.10205}.

\bibitem[{Xiao et~al.(2023)Xiao, Liu, Zhang, and Muennighoff}]{bge_embedding}
Shitao Xiao, Zheng Liu, Peitian Zhang, and Niklas Muennighoff. 2023.
\newblock \href {https://arxiv.org/abs/2309.07597} {C-pack: Packaged resources to advance general chinese embedding}.
\newblock \emph{Preprint}, arXiv:2309.07597.

\bibitem[{Xue et~al.(2024)Xue, Zhang, Dong, and Tang}]{xue-etal-2024-autore}
Lilong Xue, Dan Zhang, Yuxiao Dong, and Jie Tang. 2024.
\newblock \href {https://doi.org/10.18653/v1/2024.acl-demos.20} {{A}uto{RE}: Document-level relation extraction with large language models}.
\newblock In \emph{Proceedings of the 62nd Annual Meeting of the Association for Computational Linguistics (Volume 3: System Demonstrations)}, pages 211--220, Bangkok, Thailand. Association for Computational Linguistics.

\bibitem[{Yao et~al.(2019)Yao, Ye, Li, Han, Lin, Liu, Liu, Huang, Zhou, and Sun}]{yao-etal-2019-docred}
Yuan Yao, Deming Ye, Peng Li, Xu~Han, Yankai Lin, Zhenghao Liu, Zhiyuan Liu, Lixin Huang, Jie Zhou, and Maosong Sun. 2019.
\newblock \href {https://doi.org/10.18653/v1/P19-1074} {{D}oc{RED}: A large-scale document-level relation extraction dataset}.
\newblock In \emph{Proceedings of the 57th Annual Meeting of the Association for Computational Linguistics}, pages 764--777, Florence, Italy. Association for Computational Linguistics.

\bibitem[{Yuan et~al.(2023)Yuan, Xie, and Ananiadou}]{yuan2023zero}
Chenhan Yuan, Qianqian Xie, and Sophia Ananiadou. 2023.
\newblock Zero-shot temporal relation extraction with chatgpt.
\newblock \emph{arXiv preprint arXiv:2304.05454}.

\bibitem[{Zaratiana et~al.(2024)Zaratiana, Tomeh, Holat, and Charnois}]{zaratiana-etal-2024-gliner}
Urchade Zaratiana, Nadi Tomeh, Pierre Holat, and Thierry Charnois. 2024.
\newblock \href {https://doi.org/10.18653/v1/2024.naacl-long.300} {{GL}i{NER}: Generalist model for named entity recognition using bidirectional transformer}.
\newblock In \emph{Proceedings of the 2024 Conference of the North American Chapter of the Association for Computational Linguistics: Human Language Technologies (Volume 1: Long Papers)}, pages 5364--5376, Mexico City, Mexico. Association for Computational Linguistics.

\bibitem[{Zhang et~al.(2025)Zhang, Wu, Yu, Wu, Zheng, Huang, Zhu, Peng, Zan, and Song}]{zhang-etal-2025-cadrl}
Kunli Zhang, Pengcheng Wu, Bohan Yu, Kejun Wu, Aoze Zheng, Xiyang Huang, Chenkang Zhu, Min Peng, Hongying Zan, and Yu~Song. 2025.
\newblock \href {https://aclanthology.org/2025.coling-main.551/} {{C}a{DRL}: Document-level relation extraction via context-aware differentiable rule learning}.
\newblock In \emph{Proceedings of the 31st International Conference on Computational Linguistics}, pages 8272--8284, Abu Dhabi, UAE. Association for Computational Linguistics.

\bibitem[{Zhang and Kang(2024)}]{zhang2024tpn}
Yu~Zhang and Zhao Kang. 2024.
\newblock Tpn: Transferable proto-learning network towards few-shot document-level relation extraction.
\newblock In \emph{2024 International Joint Conference on Neural Networks (IJCNN)}, pages 1--9. IEEE.

\bibitem[{Zhang et~al.(2017)Zhang, Zhong, Chen, Angeli, and Manning}]{zhang-etal-2017-position}
Yuhao Zhang, Victor Zhong, Danqi Chen, Gabor Angeli, and Christopher~D. Manning. 2017.
\newblock \href {https://doi.org/10.18653/v1/D17-1004} {Position-aware attention and supervised data improve slot filling}.
\newblock In \emph{Proceedings of the 2017 Conference on Empirical Methods in Natural Language Processing}, pages 35--45, Copenhagen, Denmark. Association for Computational Linguistics.

\bibitem[{Zhou et~al.(2021)Zhou, Huang, Ma, and Huang}]{zhou2021atlop}
Wenxuan Zhou, Kevin Huang, Tengyu Ma, and Jing Huang. 2021.
\newblock Document-level relation extraction with adaptive thresholding and localized context pooling.
\newblock In \emph{Proceedings of the AAAI Conference on Artificial Intelligence}.

\bibitem[{Zhou et~al.(2023)Zhou, Zhang, Gu, Chen, and Poon}]{zhou2023universalner}
Wenxuan Zhou, Sheng Zhang, Yu~Gu, Muhao Chen, and Hoifung Poon. 2023.
\newblock \href {https://arxiv.org/abs/2308.03279} {Universalner: Targeted distillation from large language models for open named entity recognition}.

\end{thebibliography}

\newpage
\appendix
\section{Appendix}

\subsection{Pretraining Data Generation}
\label{sec:appendix_pretraining}
\begin{figure}[h!]
\centering
\includegraphics[width=\columnwidth]{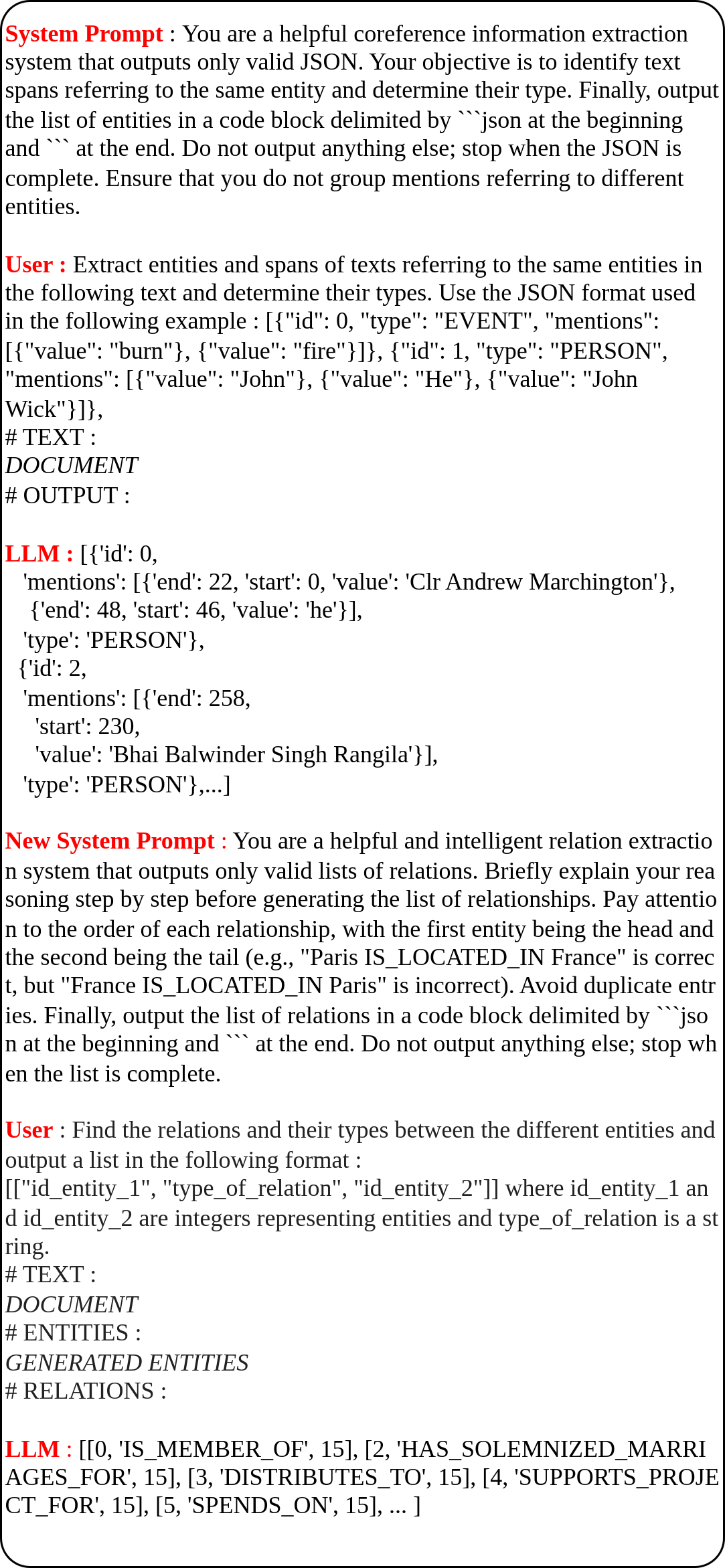}
\caption{The prompt template used for pretraining data generation with Mistral-Small.}
\label{fig:prompt_generation}
\end{figure}

\begin{figure}[ht!]
 \centering
 \includegraphics[width=\columnwidth]{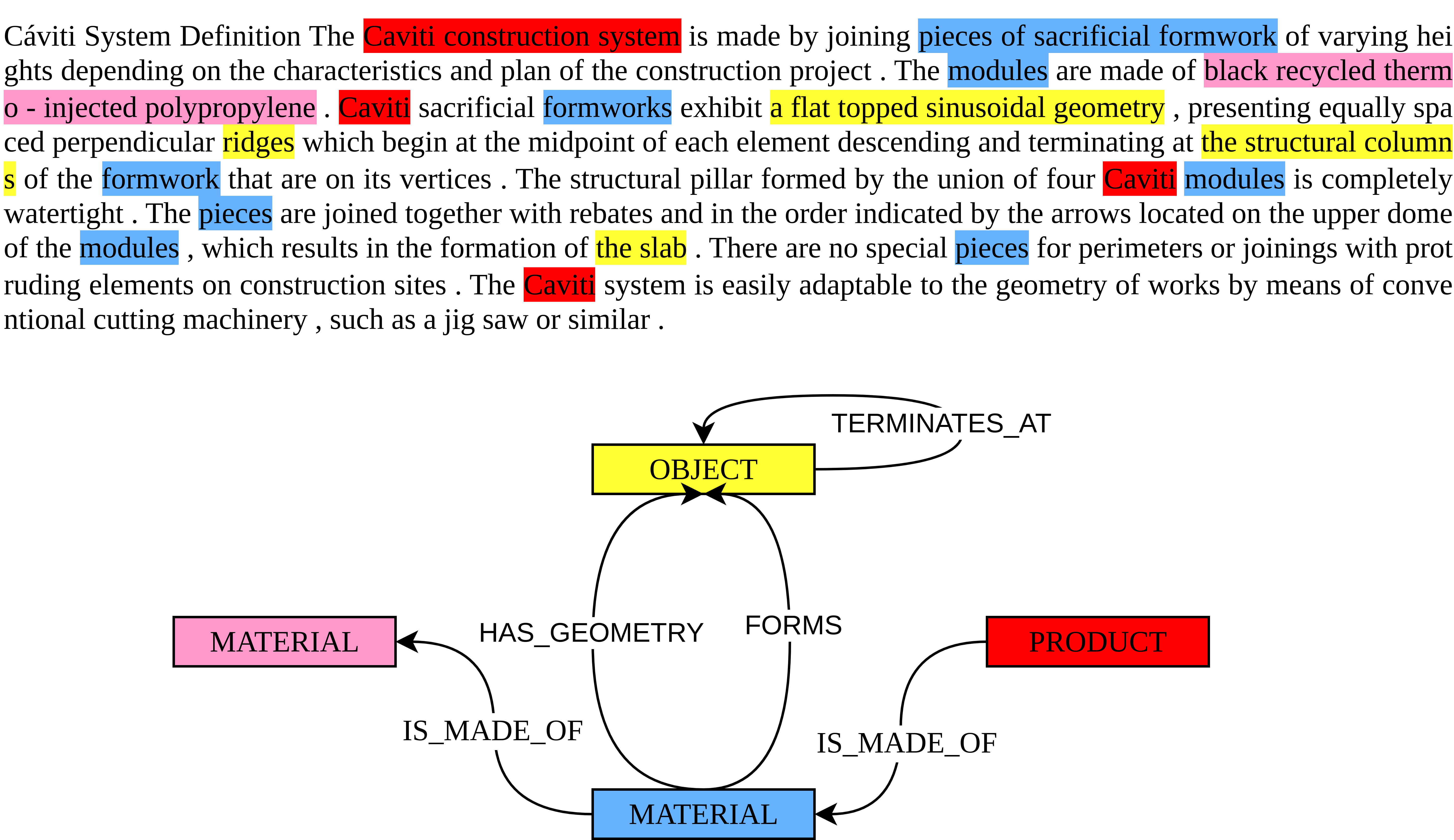}
 \caption{Example of generated data.}
 \label{fig:example_generated}
\end{figure}

The synthetic data for the pretraining phase is generated using the prompt template detailed in Figure \ref{fig:prompt_generation}. An example of generated data is provided in Figure~\ref{fig:example_generated}

\subsection{Zero-shot DocRE inference with LLMs}\label{sec:appendix_prompt_zero_shot}

\begin{figure}[H]
 \centering
 \includegraphics[width=\columnwidth]{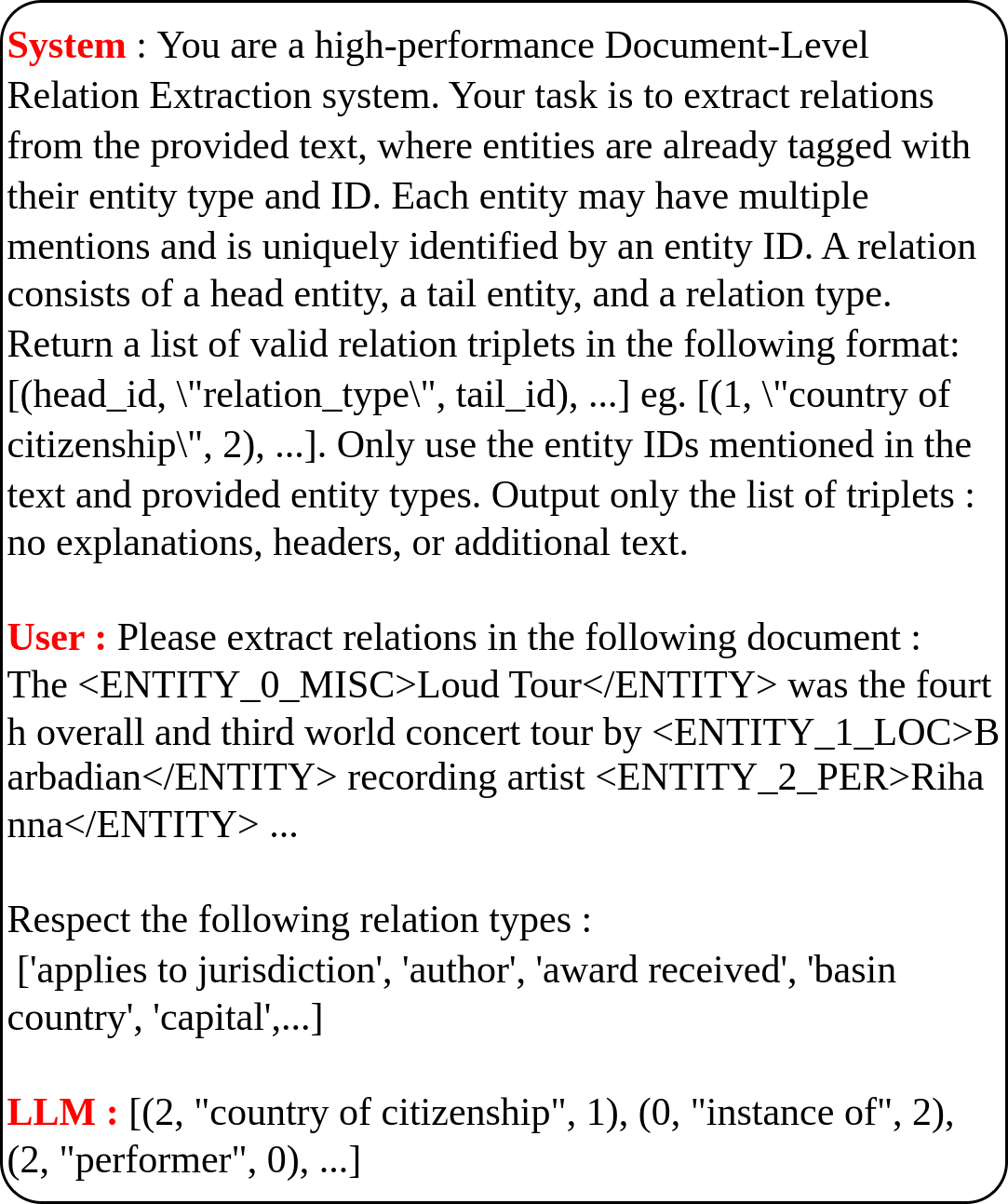}
 \caption{The prompt template used for zero-shot DocRE inference with LLMs. Entity mentions are marked with special tags to ground the model and the output format is strictly constrained to triplets to ensure parsability and reduce invalid generations.}
 \label{fig:prompt_zero_shot}
\end{figure}

\subsection{Analysis}
\label{app:analysis}
\subsubsection{LogSumExp vs. Mean Pooling}

LogSumExp (LSE) pooling serves as a smooth approximation of the max pooling operation. The operation is defined as:
$$
\text{LSE}(\mathbf{x_1,...,x_n}) = \log \left( \sum_{i=1}^{n} \exp(x_i) \right)
$$
This pooling strategy was successfully applied to document-level relation extraction by \citet{jia-etal-2019-document} and later adopted by ATLOP, where it showed slightly better performances over conventional mean pooling.

However, our empirical results, shown in Table \ref{tab:logsumexp_vs_mean}, indicate a different outcome within our architecture. For GLiDRE, conventional mean pooling outperforms LSE by nearly a full F1 point on the test set. We attribute this discrepancy to architectural differences; unlike the bilinear classifier of ATLOP, GLiDRE employs a bi-encoder framework that compares relation and type embeddings. This structural divergence suggests that optimizations are not always directly transferable between models. Given that the original ATLOP paper reported only minor gains from LSE, our findings confirm that mean pooling is a more effective and robust choice for our model.

\begin{table}[H]
\centering
\caption{Comparison of F1 scores on the test set using LogSumExp versus Mean pooling for entity mention aggregation.}
\label{tab:logsumexp_vs_mean}
\begin{tabular}{lcc}
\toprule
\textbf{Pooling Method} & \textbf{Test F1} & \textbf{Test Ign F1} \\
\midrule
LogSumExp & $76.86_{\pm 0.15}$ & $75.72_{\pm 0.20}$ \\
Mean & $\mathbf{77.83}_{\pm 0.23}$ & $\mathbf{76.80}_{\pm 0.22}$ \\
\bottomrule
\end{tabular}
\end{table}

\subsubsection{Effect of pretraining and Localized Context Pooling}

We conduct an ablation study to isolate the individual contributions of our synthetic pretraining stage and the Localized Context Pooling (noted LOP in the ATLOP model) mechanism. Two variants of our model are evaluated: one finetuned without pretraining and another without the LOP module.

The results in Table \ref{tab:pretraining_local_context} confirm that both components positively contribute to the final performance of GLiDRE. Removing the pretraining stage leads to the most significant performance decrease, with a drop of nearly $0.7$ F1 points, underscoring the effectiveness of our synthetic data generation for model initialization. Disabling LOP results in a drop of $0.2$ F1 points, which validates its role in refining relation representations. These findings justify the inclusion of both techniques in our final model architecture.

\begin{table}[H]
\centering
\caption{Ablation study on the Test set. We report F1 scores after removing the pretraining stage and the Localized Context Pooling (LOP) module.}
\label{tab:pretraining_local_context}
\begin{tabular}{lcc}
\toprule
\textbf{Configuration} & \textbf{Test F1} & \textbf{Test Ign F1} \\
\midrule
\textbf{GLiDRE} & $\mathbf{77.83}_{\pm 0.23}$ & $\mathbf{76.80}_{\pm 0.22}$ \\
\textit{w/o pretraining} & $77.15_{\pm 0.42}$ & $75.96_{\pm 0.49}$ \\
\textit{w/o LOP} & $77.61_{\pm 0.09}$ & $76.48_{\pm 0.11}$ \\
\bottomrule
\end{tabular}
\end{table}

\subsubsection{Adapting the Adaptive Threshold Loss}

ATLOP introduced an adaptive thresholding mechanism to learn a dynamic, per-relation decision boundary, thereby avoiding a suboptimal, fixed global threshold. This is achieved by adding a special "threshold" (\texttt{TH}) class to the classifier; a relation is predicted only if its logit surpasses that of the \texttt{TH} class.

Adapting this to GLiDRE is non-trivial due to our bi-encoder architecture, which lacks a fixed classifier head. To replicate the mechanism, we introduce a small multi-layer perceptron (MLP) to predict the threshold logit. The MLP input is a concatenation of the candidate relation embedding and a global context vector, formed by averaging all possible relation type embeddings. The model is then trained using the original ATLOP loss function.

To save compute resources, this experiment is conducted on the model variant without pretraining. As shown in Table \ref{tab:atloss}, this adaptation proved detrimental, degrading performance by approximately $1.4$ F1 points. We hypothesize several reasons for this negative result:
    (1) our method of computing the threshold logit via a separate MLP is fundamentally different from ATLOP's integrated classifier approach; 
    (2) the ATLOP loss may be less effective at handling the severe class imbalance in DocRED compared to the Focal Loss used in our main model;
    (3) a dynamic threshold may be unnecessary for our model. We observed that the optimal global threshold for GLiDRE consistently converges near 0.5 and further tuning per-class thresholds on the development set did not improve test set performance.

\begin{table}[H]
\centering
\caption{Comparison between standard training with Focal Loss and our adaptation of the ATLOP adaptive thresholding method. Experiments are conducted without the pretraining phase.}
\label{tab:atloss}
\begin{tabular}{lcc}
\toprule
\textbf{Configuration} & \textbf{Test F1} & \textbf{Test Ign F1} \\
\midrule
Focal Loss & $\mathbf{77.15}_{\pm 0.42}$ & $\mathbf{75.96}_{\pm 0.49}$ \\
Adaptive Threshold & $75.74_{\pm 1.05}$ & $74.84_{\pm 1.05}$ \\
\bottomrule
\end{tabular}
\end{table}

\subsection{Zero-Shot Sentence-level Relation Extraction}
\label{sec:appendix_fewrel}
To assess the generalization capabilities of our document-level model, we evaluated GLiDRE on the sentence-level zero-shot benchmarks, FewRel and Wiki-ZSL. These datasets mainly feature a single candidate entity pair per sentence, making the task closer to relation classification. Following standard protocols, we evaluate on splits where the set of $m$ test relations is disjoint from the relations seen during training.

The results are presented in Table \ref{tab:table_fewrel}. Despite being designed and pre-trained for the more complex document-level setting, GLiDRE demonstrates respectable performance. For a small number of unseen relations (m=5), our model is highly competitive, outperforming several strong baselines and GPT-4o.

However, as the number of unseen relations increases, GLiDRE performance degrades more rapidly than models specifically architected for sentence-level zero-shot relation classification, such as GLiREL and TMC-BERT. We attribute this to our model pretraining on multi-mention instances, which may make it less specialized for the sentence-level relation classification. Furthermore, its bi‑encoder design also scores each relation on its own, making it harder to tell apart similar relation types.

\begin{table*}[h!]
\centering
\begin{tabular}{clccccccccc}
\multirow{2}{*}{$m$} & \multirow{2}{*}{\textbf{Model}} & \multicolumn{3}{c}{\textbf{Wiki-ZSL}} & \multicolumn{3}{c}{\textbf{FewRel}} \\
                     &                        & P     & R     & F1    & P     & R     & F1    \\
\midrule
\multirow{7}{*}{5} 
& RelationPrompt \cite{chia-etal-2022-relationprompt}         & 70.66 & 83.75 & 76.63 & 90.15 & 88.50 & 89.30 \\
& DSP-ZRSC \cite{lv-etal-2023-dsp}               & 94.10  & 77.10  & 84.80  & 93.40  & 92.50  & 92.90  \\
& ZSRE \cite{tran2023enhancing}    & \textbf{94.50} & \textbf{96.48} & \textbf{95.46} & 96.36 & \textbf{96.68} & \textbf{96.51} \\
& MC-BERT \cite{MC-BERT-Lan-et-al}                 & 80.28 & 84.03 & 82.11 & 90.82 & 91.30 & 90.47 \\
& TMC-BERT  \cite{moller2024incorporating}             & 90.11 & 87.89 & 88.92 & 93.94 & 93.30 & 93.62 \\
& GPT-4o& 91.24& 72.07& 80.03& 96.75& 83.05&89.20\\
& GLiREL \cite{boylan-etal-2025-glirel} & 89.41& 80.67& 83.28& \textbf{96.84}& 93.41&94.20\\
&\textbf{GLiDRE} & 93.17 & 92.15 & 92.24 & 93.68 & 92.14 & 92.15 \\
\midrule
\multirow{7}{*}{10} 
& RelationPrompt \cite{chia-etal-2022-relationprompt}         & 68.51 & 74.76 & 71.50 & 80.33 & 79.62 & 79.96 \\
& DSP-ZRSC \cite{lv-etal-2023-dsp}               & 80.00  & 74.00  & 76.90  & 80.70  & 88.00  & 84.20  \\
& ZSRE \cite{tran2023enhancing}     & 85.43 & \textbf{88.14} & \textbf{86.74} & 81.13 & 82.24 & 81.68 \\
& MC-BERT \cite{MC-BERT-Lan-et-al}                & 72.81 & 73.96 & 73.38 & 86.57 & 85.27 & 85.92 \\
& TMC-BERT \cite{moller2024incorporating}              & 81.21 & 81.27 & 81.23 & 84.42 & 84.99 & 85.68 \\
& GPT-4o& 77.62& 66.14& 68.35& 84.07& 58.00&66.20\\
& GLiREL \cite{boylan-etal-2025-glirel}& \textbf{89.87} & 81.56& 83.67& \textbf{91.09}& \textbf{87.42}&\textbf{87.60}\\
&\textbf{GLiDRE} & 73.98 & 73.11 & 70.89 & 86.16 & 82.92 & 81.74 \\
\midrule
\multirow{7}{*}{15} 
& RelationPrompt NG \cite{chia-etal-2022-relationprompt}      & 54.45 & 29.43 & 37.45 & 66.49 & 40.05 & 49.38 \\
& DSP-ZRSC  \cite{lv-etal-2023-dsp}              & 77.50  & 64.40  & 70.40  & 82.90  & 78.10  & 80.40  \\
& ZSRE \cite{tran2023enhancing}    & 64.68 & 65.01 & 65.30 & 66.44 & 69.29 & 67.82 \\
& MC-BERT  \cite{MC-BERT-Lan-et-al}              & 65.71 & 67.11 & 66.40 & 80.71 & 79.84 & 80.27 \\
& TMC-BERT \cite{moller2024incorporating}               & 73.62 & 74.07 & 73.77 & 82.11 & 79.93 & 81.00 \\
& GPT-4o & \textbf{81.04}& 32.06& 41.57& 84.42& 65.76&70.70\\
& GLiREL \cite{boylan-etal-2025-glirel}& 79.44& \textbf{74.81}& \textbf{73.91}& \textbf{88.14}& \textbf{84.69}&\textbf{84.48}\\
&\textbf{GLiDRE} & 67.29 & 67.67 & 64.98 & 79.37 & 76.05 & 75.50 \\
\end{tabular}
\caption{Zero-shot performance comparison on the Wiki-ZSL and FewRel datasets for a varying number of unseen relations ($m$). Baseline results are reported from their respective original publications. GPT-4o results are from \citet{boylan-etal-2025-glirel}.}
\label{tab:table_fewrel}
\end{table*}

\end{document}